\documentclass[11pt,letterpaper,logo,twocolumn]{style}
\usepackage[numbers]{natbib}
\usepackage{graphicx}
\usepackage{booktabs}
\usepackage{amsmath,amsfonts,amssymb}
\usepackage{subcaption}
\usepackage{multirow}
\usepackage{colortbl}
\usepackage{listings}
\usepackage{xparse}
\usepackage{fontawesome5}
\usepackage{threeparttable}

\graphicspath{{./}{fig/}{figures/}{plot/}{pdf/}{table/}}
\usepackage{amsthm}
\usepackage{tcolorbox}
\tcbuselibrary{skins,breakable}
\tcbuselibrary{listingsutf8}
\usepackage{titletoc}
\usepackage{pifont}
\usepackage{mathtools}
\usepackage{bbm}
\usepackage{makecell}
\usepackage{adjustbox}

\title{Omni-R1: Towards the Unified Generative Paradigm for Multimodal Reasoning}
\runningtitle{Omni-R1: Towards the Unified Generative Paradigm for Multimodal Reasoning}
\PublicDate{2026-1-14}

\author{%
  {\Authfont
    \textbf{Dongjie Cheng}\textsuperscript{1} \quad
    \textbf{Yongqi Li}\textsuperscript{1}\advisor \quad
    \textbf{Zhixin Ma}\textsuperscript{2} \quad
    \textbf{Hongru Cai}\textsuperscript{1} \\
    \Authfont
    \textbf{Yupeng Hu}\textsuperscript{3} \quad
    \textbf{Wenjie Wang}\textsuperscript{4} \quad
    \textbf{Liqiang Nie}\textsuperscript{5} \quad
    \textbf{Wenjie Li}\textsuperscript{1}
  }\\
  {\Affilfont
    \textsuperscript{1} The Hong Kong Polytechnic University \quad
    \textsuperscript{2} Singapore Management University \quad
    \textsuperscript{3} Shandong University \\
    \textsuperscript{4} University of Science and Technology of China \quad
    \textsuperscript{5} Harbin Institute of Technology (Shenzhen) \\
    \texttt{dong-jie.cheng@connect.polyu.hk, liyongqi0@gmail.com}
  }
}

\begin{document}
\begin{abstract}
Multimodal Large Language Models (MLLMs) are making significant progress in multimodal reasoning. Early approaches focus on pure text-based reasoning. More recent studies have incorporated multimodal information into the reasoning steps; however, they often follow a single task-specific reasoning pattern, which limits their generalizability across various multimodal tasks. In fact, there are numerous multimodal tasks requiring diverse reasoning skills, such as zooming in on a specific region or marking an object within an image. To address this, we propose unified generative multimodal reasoning, which unifies diverse multimodal reasoning skills by generating intermediate images during the reasoning process. We instantiate this paradigm with Omni-R1, a two-stage SFT+RL framework featuring perception alignment loss and perception reward, thereby enabling functional image generation. Additionally, we introduce Omni-R1-Zero, which eliminates the need for multimodal annotations by bootstrapping step-wise visualizations from text-only reasoning data. Empirical results show that Omni-R1 achieves unified generative reasoning across a wide range of multimodal tasks, and Omni-R1-Zero can match or even surpass Omni-R1 on average, suggesting a promising direction for generative multimodal reasoning.
\end{abstract}

\newcommand{\TitleLinks}{%
\centering
    \vspace{6pt}
    {\noindent\absfont\fontsize{11}{13}\selectfont
    \faGithub\ Project Page: \url{https://github.com/ModalityDance/Omni-R1}\par}%
}

\maketitle

\section{Introduction}
\begin{figure}[t]
    \centering
    \includegraphics[width=\linewidth]{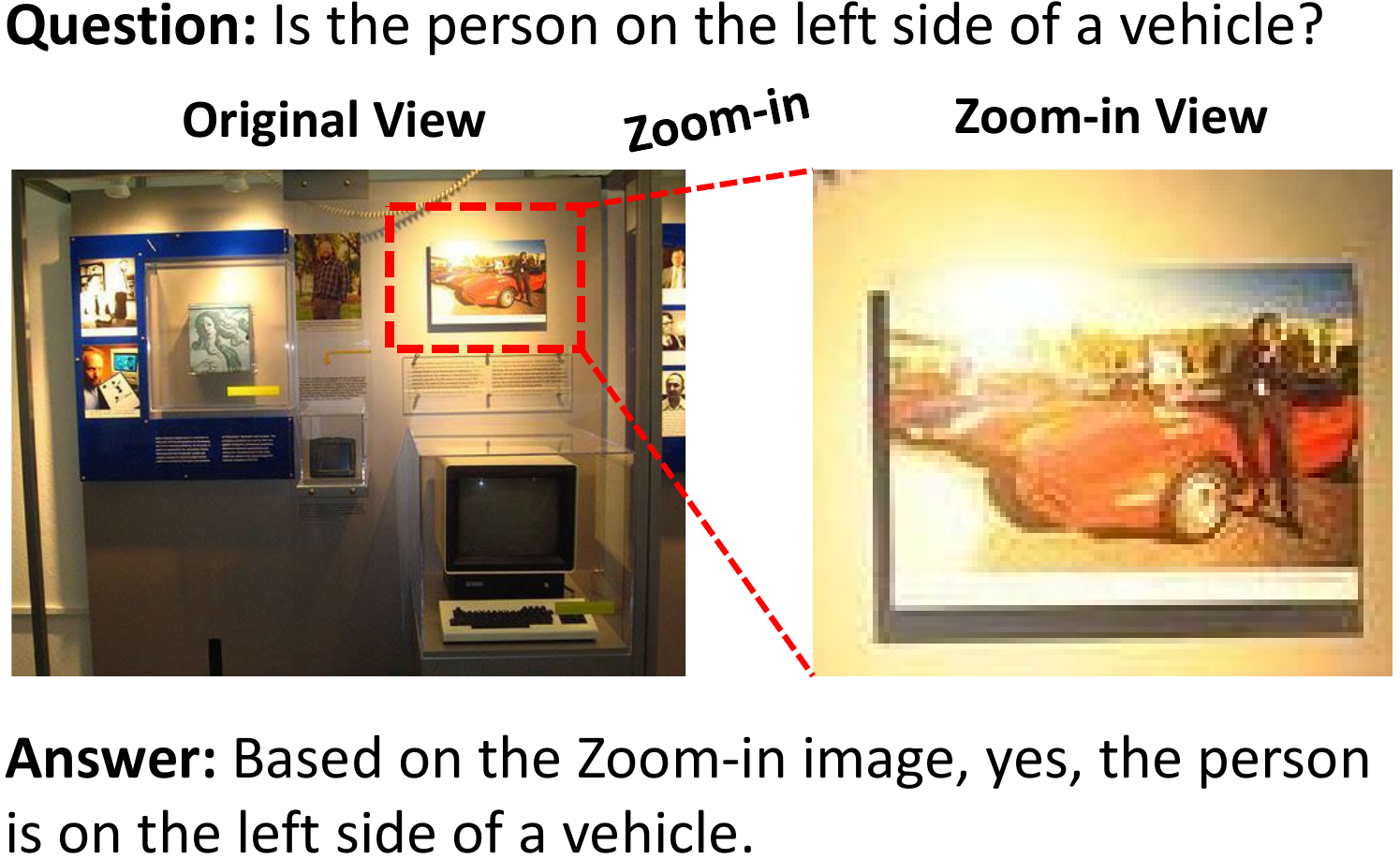}
    \caption{An example illustrating the necessity of incorporating visual information into the intermediate reasoning steps for multimodal tasks.}
    \vspace{-1em}
    \label{fig:intro_case}
\end{figure}

Multimodal Large Language Models (MLLMs) are evolving from basic perception to more advanced reasoning capabilities \citep{meng2025mmeureka}. Early multimodal reasoning methods primarily focus on pure text-based reasoning, where MLLMs perform textual reasoning before generating the final answer. More recently, some studies~\citep{hu2024visual} highlight that the intermediate reasoning steps also require the involvement of visual information. For example, as illustrated in Figure \ref{fig:intro_case}, a spatial relation problem can be more easily solved by zooming in on the critical region during the intermediate reasoning process.

\begin{figure*}[t]
    \centering
    \includegraphics[width=\linewidth]{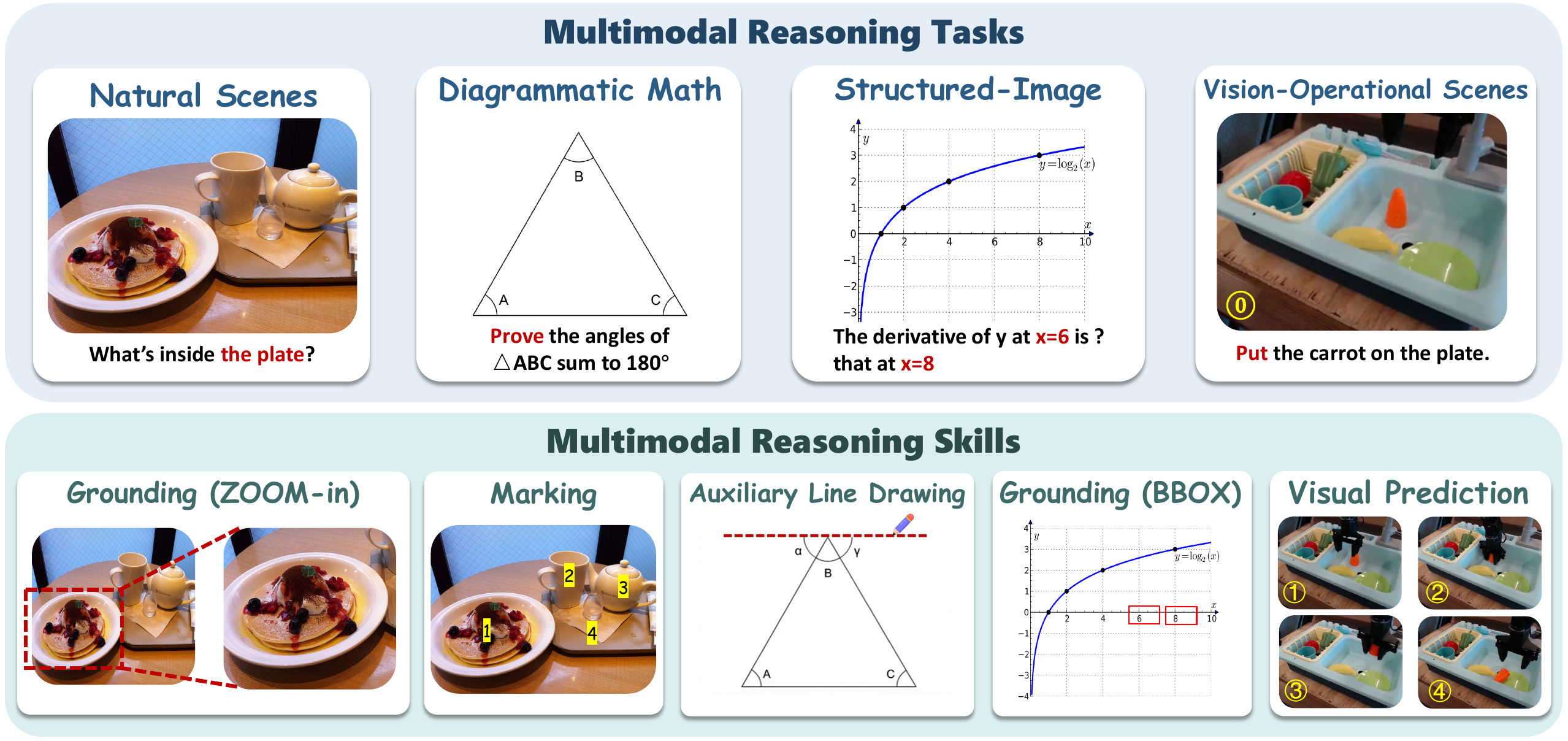}
    \vspace{-1em}
    \caption{The illustration of various multimodal tasks and the corresponding diverse multimodal reasoning skills, with multimodal tasks shown in the top row and the required reasoning skills summarized in the bottom row.}
    \vspace{-1em}
    \label{fig:tasks-skills}
\end{figure*}

There have been some interleaved-modal reasoning methods \citep{zheng2025deepeyes,li2025mvot,chen2025mintcot,chern2025twgi} to incorporate multimodal information into the reasoning steps. For example, in DeepEyes \citep{zheng2025deepeyes}, the method utilizes external tools to zoom in on a specific region of an image, enabling the model to focus on particular areas of the input image for VQA tasks. In MVoT \citep{li2025mvot}, the approach involves imagining possible paths by visualizing intermediate states to solve spatial reasoning tasks. Despite the effectiveness, current interleaved-modal reasoning methods still rely on a single specific multimodal reasoning pattern to address a particular multimodal task.

To address the above issue, we propose to develop a general and unified multimodal reasoning paradigm capable of handling a broad range of multimodal tasks. 1) First, there are numerous multimodal tasks that require diverse reasoning skills in practice. As shown in Figure \ref{fig:tasks-skills}, skills such as zoom-in and grounding are essential for general VQA tasks, while visual prediction skills are clearly beneficial for visual-operational scenes. 2) Second, it is necessary to facilitate various multimodal reasoning skills within a single MLLM in a unified paradigm. Inspired by the recent development of Omni models~\citep{team2024chameleon, chern2024anole} that exhibit multimodal generation capabilities, we propose that all multimodal reasoning skills can be unified and internalized within a generative paradigm, i.e., \textbf{unified generative multimodal reasoning}. For example, zoom-in can be realized by generating a magnified view of a specific image region, while grounding can be achieved by generating annotated bounding boxes over the objects in the original image.

Despite this promising vision of the generative multimodal reasoning paradigm, putting it into practice remains challenging due to the following aspects. 1) Functional image generation. Unlike typical images, the images required for various reasoning skills are functional and often unnatural, such as those containing marked numbers over specific objects. This poses a significant challenge for MLLMs. 2) Costly interleaved-modal reasoning annotations. The training of generative multimodal reasoning requires step-by-step annotations of the interleaved reasoning process. Such interleaved data is costly and rarely available at scale.

In this work, we propose Omni-R1, a framework that unifies various multimodal reasoning skills in a generative manner without relying on extensive supervision. Specifically, we introduce standard Omni-R1, which requires a two-stage optimization procedure with supervised fine-tuning (SFT) and reinforcement learning (RL) with a small amount of reasoning annotations. To address the issue of functional image generation, we introduce a perception alignment loss for SFT and a perception-calibrated reward for RL, which ensures that the model's optimization is supervised by visual perception. Furthermore, we present Omni-R1-Zero, which eliminates the need for costly interleaved-modal reasoning annotations. Instead, it introduces a bootstrapping step-wise visualization method to synthesize interleaved-modal reasoning data. Empirical results suggest that Omni-R1 enables unified generative reasoning across various multimodal reasoning tasks, while Omni-R1-Zero even surpasses Omni-R1 to some extent.

The contributions are summarized as follows:
\begin{itemize}
\item We identify a range of distinctive multimodal reasoning skills for multimodal tasks, and propose a generative multimodal reasoning paradigm to unify them within a single MLLM.

\item We introduce Omni-R1, which facilitates the unified multimodal reasoning paradigm and stabilizes functional image generation via the designed perception alignment loss and perception-calibrated reward.

\item We present Omni-R1-Zero, which eliminates the need for interleaved-modal reasoning annotations while achieving performance comparable to models trained with supervised annotations.

\end{itemize}
\section{Related Work}
\subsection{Text-based Multimodal Reasoning}
Prior work focuses on text-based multimodal reasoning, where images serve as context, and the model generates textual rationales and the final answer.
KAM-CoT~\citep{mondal2024kam} augmented multimodal chain-of-thought with structured external knowledge. More recently, reinforcement learning has been proven to be effective in multimodal reasoning. MM-EUREKA~\citep{meng2025mmeureka} trained MLLMs with rule-based, verifiable rewards, enhancing their reasoning abilities in math tasks. Visual-Thinker-R1-Zero~\citep{zhou2025r1} further applied such rule-based reinforcement learning directly to a non-SFT MLLM on math data, achieving improved visual reasoning performance.

\subsection{Interleaved-modal Reasoning}
ICOT~\citep{gao2025interleaved} introduced the concept of \emph{interleaved-modal reasoning}, where the model augments its trajectory by interleaving additional multimodal evidence. DeepEyes~\citep{zheng2025deepeyes} used tools to acquire and append new multimodal observations during reasoning, enabling fine-grained improvements through reinforcement learning. More recently, interleaving has been realized within a single omni-model as interleaved generation. MVoT~\citep{li2025mvot} trained an omni-MLLM to generate image-text interleaved rationale in a spatial reasoning task. To support such behaviors with supervision, Zebra-CoT~\citep{li2025zebra} released a 182K interleaved text-image reasoning dataset for the exploration of interleaved-modal reasoning, especially in omni-models. 
There is another work with the same name, Omni-R1~\citep{Zhong2025OmniR1RL}, which uses RL to optimize the selection of informative keyframes and pixel-level grounding for video segmentation. 
\section{Omni-Bench}
\label{sec:omni-bench}
To better evaluate MLLM's multimodal reasoning capabilities, we construct Omni-Bench, which consists of various multimodal tasks that require diverse multimodal reasoning skills.

\paragraph{Uni-Tasks.}
Uni-Tasks characterize the multimodal reasoning scenarios included in Omni-Bench.
\textit{Natural-Scene Perception} focuses on natural-scene images and requires evidence localization for answering.
\textit{Diagrammatic Math} involves diagram-grounded visual arithmetic and geometric reasoning.
\textit{Structured-Image} targets structured image inputs, such as figures and charts, that combine text and graphics.
\textit{Vision-Operational Scenes}, such as visual games and embodied planning, require complex visual operations or even predicting the transition of scene states.
For each Uni-Task, we randomly sampled data from established datasets to form a slice.
Table~\ref{tab:omnibench_data} summarizes the source datasets and the number of prompts in each slice.

\paragraph{Uni-Skills.}
Across these scenarios, generative multimodal reasoning should require a set of recurring visual skills for functional image generation.
We summarize them as four Uni-Skills:
\textit{Grounding} localizes task-relevant evidence and may zoom in for closer inspection when needed.
\textit{Auxiliary line drawing} draws helper lines to make geometric relations or alignment constraints explicit.
\textit{Marking} highlights or enumerates relevant instances so they can be referred to unambiguously.
\textit{Visual prediction} anticipates the next visual state by performing a one-step transition.

\begin{table}[t]
\centering
\scriptsize
\setlength{\tabcolsep}{3pt}
\renewcommand{\arraystretch}{0.95}
\resizebox{\columnwidth}{!}{%
\begin{tabular}{l l r}
\toprule
Task & Source & \ \#Samples \\
\midrule
Natural-Scene Perception & V$^\ast$~\citep{wu2024vstar} & 100 \\
\midrule
\multirow{2}{*}{Structured-Image} & ArxivQA~\citep{li-etal-2024-multimodal-arxiv} & 100 \\
                                 & ChartQA~\citep{masry2022chartqa} & 100 \\
\midrule
\multirow{2}{*}{Diagrammatic Math} & Geometry3k~\citep{lu2021inter} & 100 \\
                                  & MathVista~\citep{lu2024mathvista} & 100 \\
\midrule
Vision-Operational Scenes & ViC-Bench~\citep{wu2025vicbench} & 300 \\
\bottomrule
\end{tabular}
}
\caption{The composition of Omni-Bench. A total of 800 samples span four Uni-Tasks.}
\vspace{-1em}
\label{tab:omnibench_data}
\end{table}

\paragraph{Evaluation.}
For evaluation, we extract the final answer span $Ans$ and use an LLM judge~\citep{zheng2023mtbench,raina2024llmjudge} with a fixed evaluation template to compare it against the gold reference, producing a binary correctness decision. Here, the LLM judge acts only as a deterministic answer-matching checker rather than an open-ended evaluator.
\begin{figure*}[t]
  \centering
  \includegraphics[width=1\textwidth]{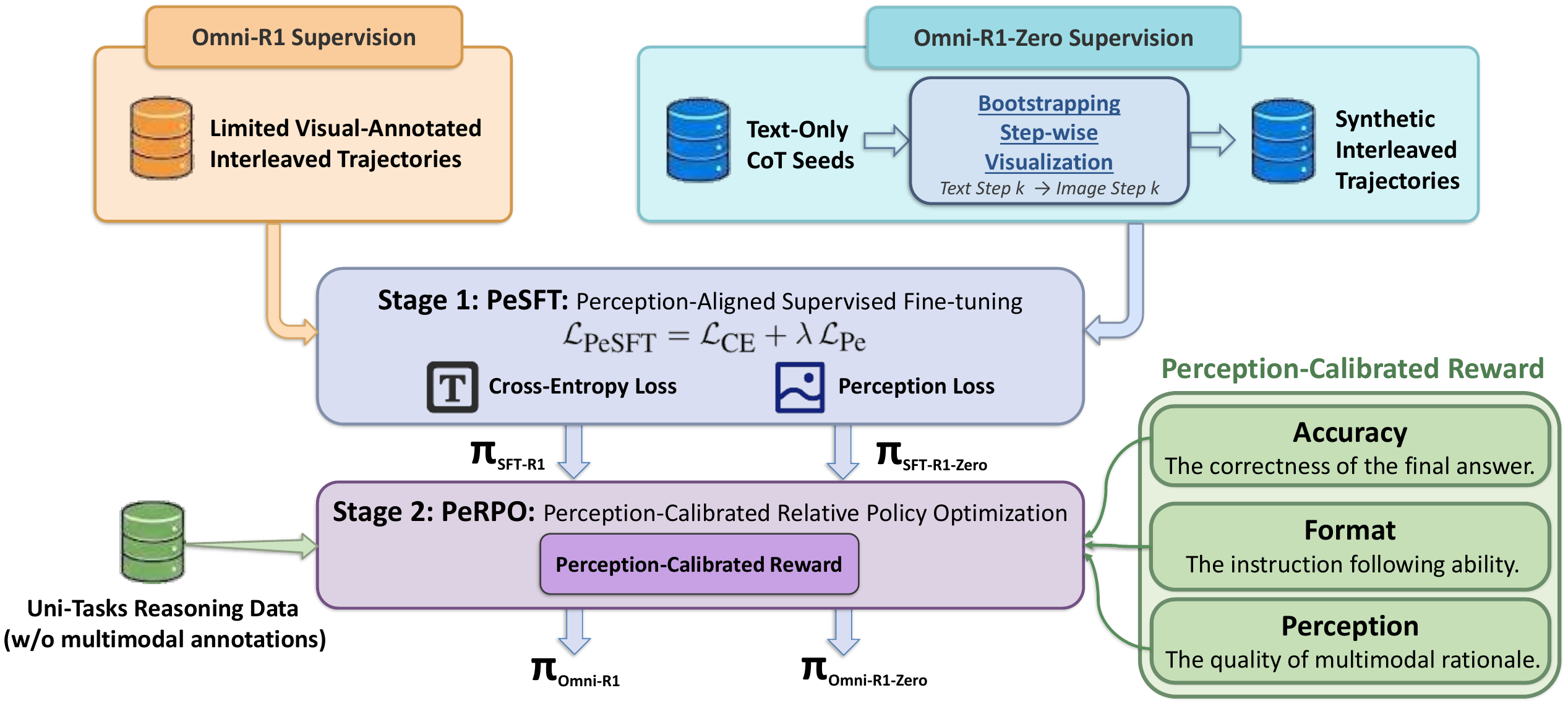}
    \caption{
      \textbf{Training pipeline for Omni-R1 and Omni-R1-Zero.}
      \textbf{(Top)} Data initialization uses either limited human annotations (orange) or synthetic bootstrapped trajectories (blue).
      \textbf{(Middle)} Stage 1 (PeSFT) performs supervised fine-tuning with joint cross-entropy and perception losses.
      \textbf{(Bottom)} Stage 2 (PeRPO) refines the policy using unified tasks without multimodal annotation (left) and a composite reward (Accuracy, Format, Perception) for final alignment.
      }
      \vspace{-1.5em}
  \label{fig:training_frameworks}
\end{figure*}

\section{Method}
\label{sec:method}

As shown in Figure~\ref{fig:training_frameworks}, we present two training frameworks, \textbf{Omni-R1} and \textbf{Omni-R1-Zero}. Both aim to learn generative multimodal reasoning while unifying diverse reasoning skills and remaining effective under limited or even zero interleaved supervision.

\subsection{Omni-R1}
\label{subsec:omnir1}

Omni-R1 optimizes the policy in two stages: 1) \textbf{Perception-Aligned Supervised Fine-Tuning (PeSFT).} This stage learns the interleaved reasoning format and Uni-Skills prediction using cross-entropy loss, and optimizes the image tokens representation with the perception loss. 2) \textbf{Perception-Calibrated Relative Policy Optimization (PeRPO).} It extends training to Uni-Tasks without multimodal annotations by refining the policy model with a perception-calibrated reward.

\subsubsection{PeSFT}
We denote the multimodal reasoning token trajectory for input $x$ as $y_{1:T}$, where $T$ indicates the token count. The output $y$ adheres to a unified template across various scenarios, as detailed in Appendix~\ref{app:problem-action}. PeSFT optimizes a cross-entropy loss to enforce this unified generation format, while applying a perception loss to align the generation of visual tokens.

\paragraph{Cross-Entropy loss.}
We apply standard cross-entropy loss over all tokens in the generative reasoning trajectory,
\begin{equation}
\small
\label{eq:ce}
\mathcal{L}_{\text{CE}}
=
\mathbb{E}\!\Big[-\sum_{t=1}^{T}\log \pi_\theta(y_t\mid x,y_{<t})\Big],
\end{equation}
which trains the model to follow or imitate the ideal reasoning process.
Overall, $\mathcal{L}_{\text{CE}}$ encourages the model to reproduce the interleaved generative reasoning format of the trajectory.
For image-token segments, it also serves as local supervision for Uni-Skills, guiding the model to generate the desired functional images given the current context.

\paragraph{Perception loss.}
We utilize a visual codebook $\mathbf{E} \in \mathbb{R}^{K \times D}$ as the reference for image tokens within the unified template, where $K$ denotes the codebook size and $D$ the embedding dimension. Let $t \in \Omega$ be the index of an image token in $y$, and $c_t \in \{1, \dots, K\}$ be its corresponding ground-truth index in $\mathbf{E}$. We define the perception loss as:
\begin{equation}
\small
\label{eq:perc}
\mathcal{L}_{\text{Pe}}
=
\frac{1}{|\Omega|}
\sum_{t\in\Omega}
\big\|W h_t - \mathbf{E}[c_t]\big\|_2^2.
\end{equation}
where $h_t \in \mathbb{R}^{H}$ is the final-layer hidden state and $W \in \mathbb{R}^{D \times H}$ is a linear projection. With $\mathbf{E}$ frozen, this loss aligns hidden states with the codebook’s geometry, acting as a perceptual prior to stabilize autoregressive image-token generation.

\paragraph{Overall training objective.} Putting the two terms together, the overall objective is
\begin{equation}
\small
\label{eq:pesft}
\mathcal{L}_{\text{PeSFT}}
=
\mathcal{L}_{\text{CE}}
+
\lambda\,\mathcal{L}_{\text{Pe}},
\qquad
\lambda=1.
\end{equation}
where we weight the perception term by a coefficient $\lambda$ and set $\lambda=1$ in all experiments.

\subsubsection{PeRPO}
PeRPO is built upon a group-relative RL objective and aims to optimize long, interleaved multimodal sequences. It introduces a perception-calibrated reward to more accurately align RL feedback with the quality of intermediate functional image generations. 

We formulate PeRPO as an optimization over the policy model’s ($\pi_{\theta}$) autoregressive trajectory generation, integrating a perception-calibrated reward $\mathcal{R}$ with a group-relative PPO objective.

\paragraph{Perception-calibrated reward.} We define the reward $\mathcal{R}$ as a weighted sum of three components, $\mathcal{R}_\mathrm{Acc}$, $\mathcal{R}_\mathrm{Fmt}$, and $\mathcal{R}_\mathrm{Pe}$.

\noindent\textbf{Accuracy ($\mathcal{R}_\mathrm{Acc}$).}
We compute an accuracy reward by parsing the final answer and comparing it to the ground truth with a rule-based verifier that checks numeric correctness, symbolic/textual equivalence, and domain-specific patterns. The accuracy reward is set to a value in $[0,1]$ based on the degree of match to the ground truth. Implementation details of the verifier are provided in Appendix~\ref{app:verifier}.

\noindent\textbf{Format ($\mathcal{R}_\mathrm{Fmt}$).}
We set $\mathcal{R}_\mathrm{Fmt}=1$ if the generated trajectory contains a well-formed reasoning part and a final answer part that can be reliably parsed by the verifier, and $\mathcal{R}_\mathrm{Fmt}=0$ otherwise.

\noindent\textbf{Perception ($\mathcal{R}_\mathrm{Pe}$).}
We measure perceptual coherence of intermediate image generations via 2D Total Variation (TV) on codebook embeddings.
Let $\mathbf{E}\in\mathbb{R}^{K\times D}$ be a frozen visual codebook.
Given a visual token index $c_t\in\{1,\ldots,K\}$, where $K$ is the codebook size and $D$ is the codebook embedding dimension, we retrieve its embedding by
\begin{equation}
\small
\label{eq:code-embed}
\mathbf{e}_t \;=\; \mathbf{E}[c_t] \;\in\; \mathbb{R}^{D}.
\end{equation}
The 1-D image-token segment with $N=H_qW_q$ tokens, where $H_{q}$ and $W_{q}$ is the height/width of the quantized image grid in VQVAE of the model, is reshaped into a $H_q\times W_q$ grid. The embedding matrix of this image can be written as
\begin{equation}
\small
\label{eq:z-shape}
\mathbf{Z}\;\in\;\mathbb{R}^{H_q\times W_q\times D}.
\end{equation}
We define the index sets $\mathcal{N}_{\mathrm{h}}$ and $\mathcal{N}_{\mathrm{v}}$ as
\begin{equation}
\small
\begin{aligned}
\mathcal{N}_{\mathrm{h}} &= \{(i,j)\mid 1\le i\le H_q,\ 1\le j< W_q\},\\
\mathcal{N}_{\mathrm{v}} &= \{(i,j)\mid 1\le i< H_q,\ 1\le j\le W_q\}.
\end{aligned}
\end{equation}
Then
\begin{equation}
\small
E_{\mathrm{h}}=\frac{1}{|\mathcal{N}_{\mathrm{h}}|}\sum_{(i,j)\in \mathcal{N}_{\mathrm{h}}}
\|\mathbf{Z}_{i,j+1}-\mathbf{Z}_{i,j}\|_2^2,
\end{equation}
and
\begin{equation}
\small
E_{\mathrm{v}}=\frac{1}{|\mathcal{N}_{\mathrm{v}}|}\sum_{(i,j)\in \mathcal{N}_{\mathrm{v}}}
\|\mathbf{Z}_{i+1,j}-\mathbf{Z}_{i,j}\|_2^2.
\end{equation}
Finally, we compute the 2-D TV energy as
\begin{equation}
\small
\label{eq:tv2d}
E_{\mathrm{2D}}=\frac{1}{D}\Big(E_{\mathrm{h}}+E_{\mathrm{v}}\Big).
\end{equation}
We map energy to a normalized score
\begin{equation}
\small
\label{eq:emap}
s_r=\frac{1}{1+E_{\mathrm{2D}}/\tau},
\end{equation}
where $\tau>0$ controls sensitivity. Let $\mathcal{S}_{\text{seg}}$ be the set of image-token segments in a trajectory.
We aggregate segment scores by
\begin{equation}
\small
\label{eq:pe-agg}
\mathcal{R}_\mathrm{Pe}=\frac{1}{|\mathcal{S}_{\text{seg}}|}\sum_{r\in\mathcal{S}_{\text{seg}}} s_r.
\end{equation}

We combine the three reward components as
\begin{equation}
\small
\label{eq:reward}
\mathcal{R}=\alpha\,\mathcal{R}_\mathrm{Acc}+\beta\,\mathcal{R}_\mathrm{Fmt}+\gamma\,\mathcal{R}_\mathrm{Pe},
\end{equation}
where $\alpha,\beta,\gamma$ balance answer accuracy, format compliance, and perceptual consistency.
We use $\mathcal{R}$ as the scalar reward $r_i$ for each candidate $i$ when computing group-relative advantages.

\paragraph{Policy optimization.}
In each optimization step, we sample a set of reasoning trajectory candidates $\mathcal{G}=\{y^{(i)}\}$ for each prompt $x$ using the behavioral policy $\pi_{\text{old}}$. Each $y^{(i)}$ is an interleaved multimodal sequence of length $T_i$ and is assigned a scalar reward $r^{(i)}$ according to Eq. \ref{eq:reward}. To facilitate meaningful gradient updates, we filter the samples to retain only mixed-outcome (non-degenerate) groups (i.e., $\sigma_{\mathcal{G}}>0$) before updating the policy network to $\pi_{\theta}$. Subsequently, we compute the group-relative advantage based on the reward:

\begin{equation}
\label{eq:adv_policy}
A^{(i)}=\frac{r^{(i)}-\mu_{\mathcal{G}}}{\sigma_{\mathcal{G}}+\delta},
\end{equation}
where $\mu_{\mathcal{G}}$ and $\sigma_{\mathcal{G}}$ denote the mean and standard deviation of the rewards
$\{r^{(j)}\}_{j\in\mathcal{G}}$, and $\delta>0$ is a small constant for numerical stability.
Let $m_{i,t}\in\{0,1\}$ be the response mask for candidate $y^{(i)}$, where $t\in\{1,\ldots,T_i\}$ indexes token positions and $m_{i,t}=1$ indicates a response position to be optimized.
We denote the number of optimized (response) positions by $L_i=\sum_{t=1}^{T_i} m_{i,t}$.

Using the computed advantages, we maximize a PPO-clip objective with a KL regularizer to a fixed reference policy $\pi_{\text{ref}}$:
\begin{align}
\resizebox{0.9\columnwidth}{!}{$\displaystyle
\begin{aligned}
\mathcal{J}(\theta)
&=\mathbb{E}_{x,\{y^{(i)}\}\sim\pi_{\text{old}}}\Bigg[
   \frac{1}{|\mathcal{G}|}\sum_{i\in\mathcal{G}}\frac{1}{L_i}\sum_{t=1}^{T_i} m_{i,t}\,
   \Bigg(
   \mathcal{L}^{(i,t)}_{\text{clip}}(\theta)\\
&\qquad\qquad\qquad\qquad\qquad
   -\beta_{\mathrm{KL}}\,D^{(i,t)}_{\mathrm{KL}}(\pi_\theta,\pi_{\text{ref}})
   \Bigg)
\Bigg].
\end{aligned}
$}
\end{align}

\noindent Here $\mathcal{L}^{(i,t)}_{\text{clip}}(\theta)$ is the PPO clipped surrogate loss,
\begin{equation}
\resizebox{0.85\columnwidth}{!}{$\displaystyle
\mathcal{L}^{(i,t)}_{\text{clip}}(\theta)
=\min\Big\{
   \rho^{(i,t)}(\theta)A^{(i)},\ 
   \operatorname{clip}\!\big(\rho^{(i,t)}(\theta),l,u\big)\,A^{(i)}
\Big\}.
$}
\end{equation}
\noindent with $l=1-\varepsilon_{\text{low}}$, $u=1+\varepsilon_{\text{high}}$, $\beta_{\mathrm{KL}}>0$, and $\mathrm{clip}(z,l,u)=\min(\max(z,l),u)$.
For the autoregressive prefix $y^{(i)}_{<t}=(y^{(i)}_1,\ldots,y^{(i)}_{t-1})$, we define the token-level likelihood ratio and KL term as
\begin{equation}
\resizebox{0.7\columnwidth}{!}{$\displaystyle
\label{eq:rho_kl_defs}
\begin{aligned}
\rho^{(i,t)}(\theta)
&=\frac{\pi_\theta\!\left(y^{(i)}_t \mid x, y^{(i)}_{<t}\right)}
        {\pi_{\text{old}}\!\left(y^{(i)}_t \mid x, y^{(i)}_{<t}\right)},\\
D^{(i,t)}_{\mathrm{KL}}(\pi_\theta,\pi_{\text{ref}})
&=\log\frac{\pi_\theta\!\left(y^{(i)}_t \mid x, y^{(i)}_{<t}\right)}
               {\pi_{\text{ref}}\!\left(y^{(i)}_t \mid x, y^{(i)}_{<t}\right)}.
\end{aligned}
$}
\end{equation}

\begin{table*}[!t]
\small
\centering
\setlength{\tabcolsep}{4pt}
\resizebox{1.0\linewidth}{!}{%
\begin{tabular}{llcccccc}
\toprule
Backbone & Model & Natural & Structured & Diagrammatic & Vision-Op. & Average & $\Delta$Avg (\%) \\
\toprule
\multirow{4}{*}{Qwen-2.5-VL-7B~\citep{bai2025qwen25vltechnicalreport}}
& Qwen-2.5-VL (Base)~\citep{bai2025qwen25vltechnicalreport} & \underline{0.710} & \underline{0.657} & 0.388 & 0.063 & 0.374 & - \\
& MM-Eureka~\citep{meng2025mmeureka} & 0.670 & 0.608 & \underline{0.489} & \textbf{0.103} & \underline{0.397} & +6.1 \\
& DeepEyes~\citep{zheng2025deepeyes} & \textbf{0.780} & 0.578 & 0.433 & 0.080 & 0.380 & +1.7 \\
& MINT-CoT~\citep{chen2025mintcot} & 0.580 & 0.647 & 0.488 & 0.017 & 0.363 & -3.0 \\
\midrule

\multirow{9}{*}{Anole-7B~\citep{chern2024anole}}
& Anole (Base)~\citep{chern2024anole} & 0.180 & 0.140 & 0.055 & 0.027 & 0.081 & - \\
& Zebra-CoT~\citep{li2025zebra} & 0.398 & 0.130 & 0.073 & 0.077 & 0.129 & +59.2 \\
\cmidrule(lr){2-8}
& \multicolumn{7}{c}{\textbf{\textit{Omni-R1}}} \\
\cmidrule(lr){2-8}
& \textbf{Omni-R1-M} & 0.440 & 0.135 & 0.115 & 0.093 & 0.153 & +87.7 \\
& \textbf{Omni-R1-L} & 0.420 & 0.175 & 0.090 & 0.090 & 0.153 & +87.7 \\
\cmidrule(lr){2-8}
& \multicolumn{7}{c}{\textbf{\textit{Omni-R1-Zero}}} \\
\cmidrule(lr){2-8}
& \textbf{Omni-R1-Zero-T} & 0.450 & 0.165 & 0.095 & 0.087 & 0.154 & +89.2 \\
& \textbf{Omni-R1-Zero-S} & 0.340 & 0.145 & 0.090 & \underline{0.097} & 0.138 & +69.2 \\
& \textbf{Omni-R1-Zero-M} & 0.410 & 0.170 & 0.120 & 0.093 & 0.159 & +95.4 \\
\midrule

\multirow{4}{*}{Bagel-7B~\citep{deng2025emergingpropertiesunifiedmultimodal}}
& Bagel (Text CoT, Base)~\citep{deng2025emergingpropertiesunifiedmultimodal} & 0.440 & 0.562 & 0.317 & 0.030 & 0.286 & - \\
& Zebra-CoT~\citep{li2025zebra} & 0.500 & 0.612 & 0.383 & 0.060 & 0.333 & +16.5 \\
\cmidrule(lr){2-8}
& \multicolumn{7}{c}{\textbf{\textit{Omni-R1}}} \\
\cmidrule(lr){2-8}
& \textbf{Omni-R1-M} & 0.560 & \textbf{0.706} & \textbf{0.498} & 0.083 & \textbf{0.402} & +40.7 \\

\bottomrule
\end{tabular}
}
\caption{Results on Omni-Bench. Column headers correspond to the four Uni-Tasks: Natural (Natural-Scene Perception), Structured (Structured-Image), Diagrammatic (Diagrammatic Math), and Vision-Op.\ (Vision-Operational). Bold numbers indicate the best result in each column, and underlined numbers indicate the second-best result. \textit{Average} is computed as a sample-weighted average over slices. $\Delta$Avg (\%) is computed w.r.t.\ the base model within the same backbone group.}
\label{tab:omni_bench}
\end{table*}

\subsection{Omni-R1-Zero}
\label{subsec:omnir1zero}
We consider the setting where human-annotated interleaved multimodal rationales are unavailable, and bootstrap interleaved supervision from text-only CoT to activate generative multimodal reasoning.

\subsubsection{Bootstrapping Step-wise Visualization}
As illustrated in Figure~\ref{fig:training_frameworks}, Omni-R1-Zero starts from text-only Chain-of-Thought (CoT) seeds and constructs synthetic interleaved trajectories via step-wise visualization, generating one image by the base model itself for each reasoning step (Text Step $k$ $\rightarrow$ Image Step $k$) and assembling them with the same control-token template as Omni-R1.

These synthetic traces are not intended to be perfect supervision, but rather to teach the interleaved reasoning format and expose intermediate multimodal states.

\subsubsection{Optimization}
Omni-R1-Zero reuses the same two-stage pipeline as Omni-R1: PeSFT on the synthetic interleaved data, followed by PeRPO with the same objective described in Section~\ref{subsec:omnir1}.
This optimization improves correctness and stabilizes intermediate image generations, enabling interleaved multimodal traces to emerge without human-annotated rationales.
\section{Experiment}
\label{sec:experiment}

\subsection{Experimental Setup}

\paragraph{Benchmarks.} We evaluated Omni-R1 and Omni-R1-Zero on Omni-Bench, which covers four Uni-Tasks, namely Natural-Scene Perception, Diagrammatic Math, Structured-Image, and Vision-Operational. We additionally reported results on standard general multimodal benchmarks, including MME~\citep{fu2023mme}, MM-Vet~\citep{yu2024mmvet}, V$^\ast$-Bench~\citep{wu2024vstar}, POPE~\citep{li2023pope}, MMVP~\citep{tong2024mmvp}, and BLINK~\citep{fu2024blink}. Detailed dataset descriptions are provided in the Appendix~\ref{sec:bench_des}. 

\paragraph{Baselines.}
We compared against diverse multimodal reasoning baselines, including Zebra-CoT~\cite{li2025zebra}, Bagel (Text CoT)~\cite{deng2025emergingpropertiesunifiedmultimodal}, MM-Eureka~\cite{meng2025mmeureka}, MINT-CoT~\cite{chen2025mintcot}, and DeepEyes~\cite{zheng2025deepeyes}. These baselines cover generative multimodal reasoning, text-only, and tool-based multimodal reasoning.

\paragraph{Evaluation.}
For Omni-Bench, we followed the evaluation protocol described in Section~\ref{sec:omni-bench}. For general benchmarks, we conducted evaluation using the widely adopted VLMEvalKit~\citep{duan2024vlmevalkit}. Unless otherwise specified, we reported accuracy, and for POPE~\citep{li2023pope}, we reported the F1-score.

\paragraph{Implementation.}
We instantiate Omni-R1 on two Unified Multimodal Model (UMM) backbones, Anole~\citep{chern2024anole} and Bagel~\cite{deng2025emergingpropertiesunifiedmultimodal}, and Omni-R1-Zero on Anole~\citep{chern2024anole}. For Omni-R1, we report the results of Omni-R1-M and Omni-R1-L, which were trained with 30 and 60 PeRPO steps, respectively. As for Omni-R1-Zero, we report Omni-R1-Zero-T, Omni-R1-Zero-S, and Omni-R1-Zero-M trained with 10, 20, and 30 PeRPO steps, respectively. Data usage is described in Appendix~\ref{sec:bench_des}, and implementation details are provided in Appendix~\ref{sec:appendix-hparams}.

\begin{table*}[!t]
\centering
\setlength{\tabcolsep}{4pt}
\resizebox{\linewidth}{!}{
\begin{tabular}{llccccccccc}
\toprule
Backbone & Model & MME-P & MME-R & MM-Vet & V$^\ast$ & POPE & MMVP & BLINK & Average & $\Delta$Avg (\%) \\
\toprule

\multirow{9}{*}{Anole-7B~\citep{chern2024anole}}
& Anole (Base)~\citep{chern2024anole} & 224.55 & 54.64 & 8.95 & 21.99 & 27.74 & 39.00 & 21.99 & 22.25 & - (-41.6) \\
& Zebra-CoT~\citep{li2025zebra} & 623.70 & 231.07 & 7.20 & 24.61 & 48.88 & 37.00 & 29.04 & 38.12 & +71.3 (-) \\
\cmidrule(lr){2-11}
& \multicolumn{10}{c}{\textbf{\textit{Omni-R1}}} \\
\cmidrule(lr){2-11}
& \textbf{Omni-R1-M} & 676.67 & 259.29 & 11.70 & 35.60 & 62.91 & 45.33 & \textbf{35.61} & 46.23 & +107.8 (+21.3) \\
& \textbf{Omni-R1-L} & \textbf{775.17} & \underline{279.64} & 11.97 & 30.90 & 61.84 & \textbf{50.67} & \underline{35.25} & \underline{48.29} & +117.0 (+26.7) \\
\cmidrule(lr){2-11}
& \multicolumn{10}{c}{\textbf{\textit{Omni-R1-Zero}}} \\
\cmidrule(lr){2-11}
& \textbf{Omni-R1-Zero-T} & 642.38 & 277.86 & \underline{17.16} & 35.08 & \textbf{66.19} & 36.67 & 31.35 & 45.73 & +105.5 (+20.0) \\
& \textbf{Omni-R1-Zero-S} & \underline{699.84} & \textbf{295.71} & \textbf{22.02} & \textbf{37.61} & \underline{66.06} & \underline{48.67} & 33.04 & \textbf{50.19} & +125.5 (+31.6) \\
& \textbf{Omni-R1-Zero-M} & 631.27 & 232.86 & 14.56 & \underline{36.96} & 65.82 & 47.00 & 30.41 & 45.16 & +102.9 (+18.4) \\
\bottomrule
\end{tabular}
}
\caption{Results on general benchmarks. Bold numbers indicate the best result in each column, underlined numbers indicate the second-best result. \textit{Average} is computed by first scaling MME-P and MME-R by 0.1 and 0.25, respectively, and then averaging over all seven metrics. $\Delta$Avg (\%) is the relative change w.r.t.\ Anole (Base), and the value in parentheses is computed w.r.t.\ Zebra-CoT.}
\vspace{-1em}
\label{tab:general_bench}
\end{table*}

\begin{table}[t]
\centering
\setlength{\tabcolsep}{4pt}
\resizebox{\linewidth}{!}{%
\begin{tabular}{lccccc}
\toprule
Model & Natural & Structured & Diagrammatic & Vision-Op. & Avg. \\
\midrule
\textbf{Omni-R1-Zero} & \textbf{0.410} & \textbf{0.170} & \textbf{0.120} & \textbf{0.093} & \textbf{0.159} \\
w/o PeRPO    & 0.390 & 0.130 & 0.075 & 0.033 & 0.113 \\
w/o PeReward & 0.330 & 0.150 & 0.085 & 0.080 & 0.130 \\
\bottomrule
\end{tabular}
}
\caption{Ablation study on Omni-Bench for Omni-R1-Zero. Results are reported on four Uni-Task slices. \textit{Avg.} denotes the average over all samples.}
\vspace{-0.5em}
\label{tab:omni_ablation}
\end{table}

\subsection{Main Results}
We presented and discussed results on Omni-Bench and general multimodal benchmarks separately.

\paragraph{Analysis on Omni-Bench.}
Table~\ref{tab:omni_bench} shows that Omni-R1 and Omni-R1-Zero perform competitively across the four Uni-Task slices, with consistent improvements over the corresponding base models. Notably, Omni-R1-M (Bagel) achieves the strongest average performance, showing that our framework is competitive with text-only and tool-based reasoning baselines. Meanwhile, Omni-R1-Zero-T/S/M (Anole) consistently outperform the base model across different RL budgets, indicating that PeRPO can successfully induce generative multimodal reasoning.

\paragraph{Analysis on general benchmarks.}
Table~\ref{tab:general_bench} shows that, on the Anole~\citep{chern2024anole} backbone, both Omni-R1 and Omni-R1-Zero outperform the baselines on standard multimodal benchmarks, indicating stronger general reasoning capability beyond Omni-Bench. On perception-oriented evaluations, including MME-P~\citep{fu2023mme}, MMVP~\citep{tong2024mmvp}, and BLINK~\citep{fu2024blink}, Omni-R1 variants are generally stronger, reflecting improved visual perception with multiple reasoning skills. By contrast, Omni-R1-Zero is more competitive on the reasoning-oriented MME-R~\citep{fu2023mme} and also performs strongly on the comprehensive MM-Vet~\citep{yu2024mmvet}. It also remains stronger on POPE~\citep{li2023pope}, suggesting better evidence consistency and reduced hallucination.

\begin{figure}[t!]
  \centering
  \resizebox{\columnwidth}{!}{%
    \begin{minipage}{\columnwidth}
      \centering
      \begin{minipage}[t]{0.49\columnwidth}
        \centering
        \includegraphics[width=\linewidth]{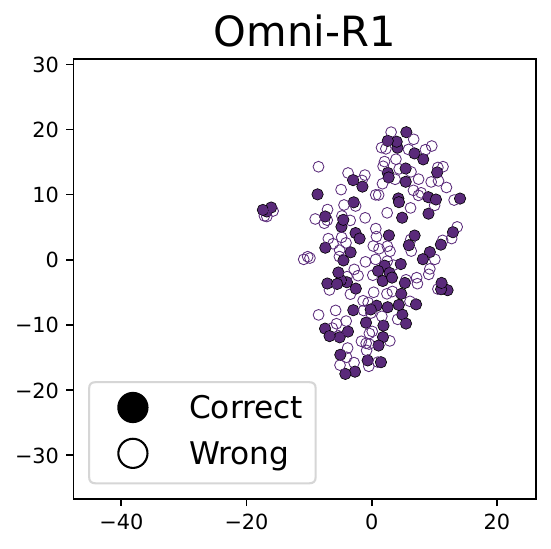}
      \end{minipage}
      \hfill
      \begin{minipage}[t]{0.49\columnwidth}
        \centering
        \includegraphics[width=\linewidth]{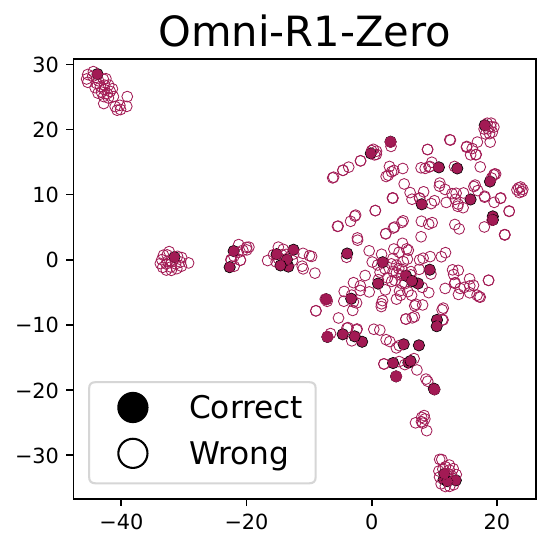}
      \end{minipage}
    \end{minipage}
  }
  \caption{The t-SNE visualization of generated images from Omni-R1 and Omni-R1-Zero, respectively. Filled markers indicate correct predictions and empty markers indicate incorrect predictions.}
  \vspace{-1em}
  \label{fig:tsne}
\end{figure}

\begin{figure*}[t!]
  \centering
  \includegraphics[width=0.85\textwidth]{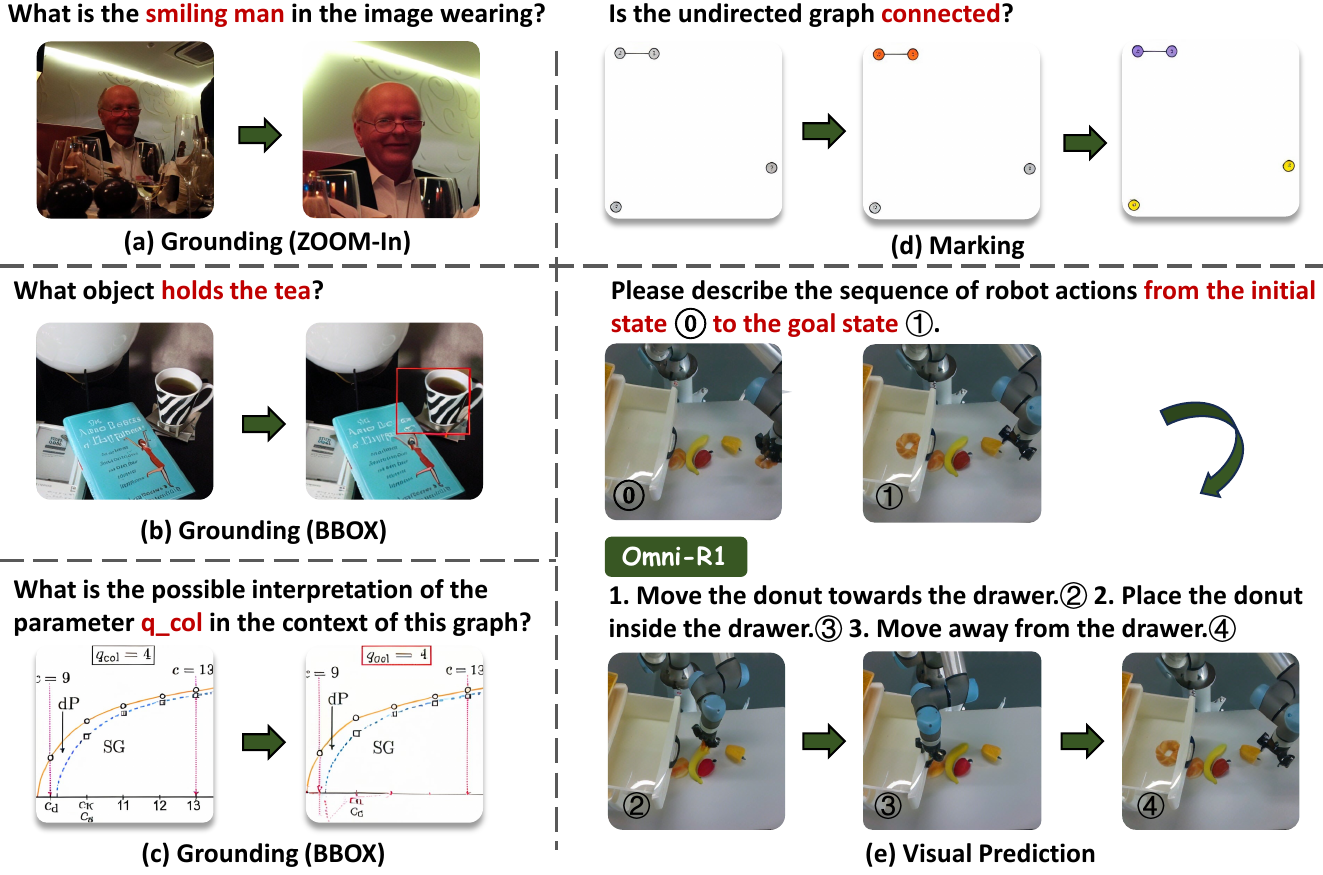}
\caption{Case studies of Omni-R1 (Anole)'s multimodal reasoning skills, including (a) Grounding (Zoom-In) for attribute recognition, (b) Grounding (BBOX) for target localization, and (c) Grounding (BBOX) for parameter localization. (d) Marking for graph connectivity verification and (e) Visual prediction for robotic manipulation sequences.}
\vspace{-0.5em}
  \label{fig:r1_case}
\end{figure*}

\subsection{In-depth Analysis}
While the main results establish consistent gains, they do not explain which components drive them or how the models behave. We therefore performed ablations and qualitative analyses on the Anole~\citep{chern2024anole} backbone to provide more insights.

\paragraph{Ablation study.}
We conducted an ablation study on Omni-R1-Zero to disentangle the effects of PeRPO and the perception-calibrated reward, by either removing PeRPO entirely (w/o PeRPO) or disabling the perception reward while keeping PeRPO (w/o PeReward). As shown in Table~\ref{tab:omni_ablation}, removing PeRPO results in the largest drop, with the regression concentrated on the Vision-Op.\ and Diagrammatic tasks. This suggests that PeRPO is crucial for developing multi-step and complex multimodal reasoning. Disabling the perception-calibrated reward yields a smaller but consistent degradation, noticeably on Natural tasks. This suggests that perception-calibrated reward improves visual evidence utilization and stabilizes policy optimization.

\paragraph{Analysis on generated multimodal patterns.}
We studied how Omni-R1 and Omni-R1-Zero shape the distribution of generated multimodal patterns differently and how these patterns relate to reasoning. In Figure~\ref{fig:tsne}, Omni-R1 concentrates into a few compact modes, whereas Omni-R1-Zero exhibits a more dispersed and multi-modal structure. In both projections, correct instances cluster more tightly than incorrect ones, suggesting that trace supervision promotes canonical and stable generations, while reward guidance encourages broader exploration that can still support correct decisions.

\begin{figure}[!t]
  \centering
  \resizebox{\columnwidth}{!}{%
    \includegraphics{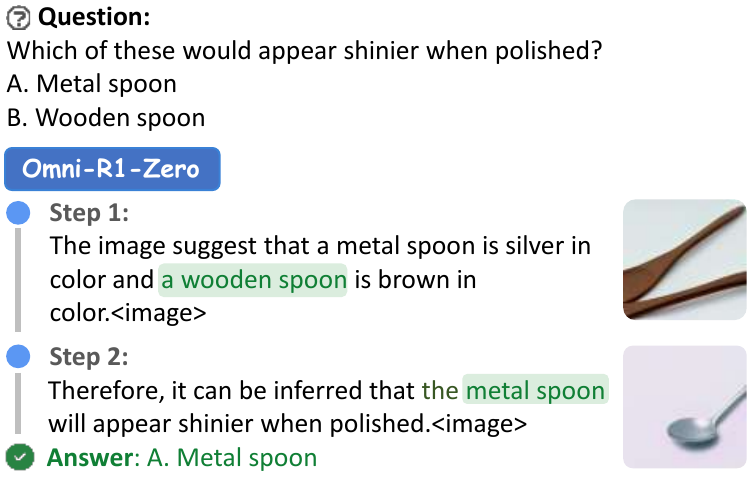}
  }
  \caption{Omni-R1-Zero's generative reasoning process on a commonsense multimodal question.}
  \vspace{-1em}
  \label{fig:zero_case}
\end{figure}

\paragraph{Case study.}
To better understand how our model behaves, we presented several examples of the reasoning trajectories produced by our models.
Figure~\ref{fig:r1_case} shows that Omni-R1 produces task-conditioned intermediate visual evidence throughout generation. This supports our claim that Uni-Skills are flexibly composed within a unified generative process.
Figure~\ref{fig:zero_case} shows that Omni-R1-Zero, despite having no supervised multimodal traces, can still generate intermediate visual evidence before getting the final answer. More complete generative multimodal reasoning cases can be found in Appendix~\ref{section:additional_res}.
\section{Conclusion and Future Work}
We revisit multimodal reasoning as generative multimodal reasoning, identifying two practical bottlenecks: functional image generation and costly interleaved-modal reasoning annotations. Omni-R1 targets the former, whereas Omni-R1-Zero targets the latter. Empirically, results on both unified and standard evaluations support these choices: Omni-R1 is particularly effective when reliable intermediate visual evidence is crucial, and Omni-R1-Zero still yields consistent gains despite limited multimodal trace supervision. A key direction is to develop scalable supervision signals for generative multimodal reasoning without annotation, thereby unlocking the full potential of zero-shot settings.

\bibliographystyle{unsrtnat} 
\bibliography{ref}

\appendix
\clearpage
\section{Problem Formulation for Generative Multimodal Reasoning}
\label{app:problem-action}

\subsection{Trajectory Formulation}
\label{app:problem-definition}

Given a multimodal input $x^{\mathrm{M}}$ (e.g., an image) and a textual input $x^{\mathrm{T}}$ (e.g., the question),
we model generative multimodal reasoning as an interleaved trajectory of textual rationales and executable visual actions.
The trajectory is represented as
\begin{equation}
\label{eq:traj}
\small
\mathcal{Y}
=\Big(
x^{\mathrm{M}},\,x^{\mathrm{T}},\,\{(rat_\ell^{\mathrm{T}},\,a_\ell,\,rat_\ell^{\mathrm{M}})\}_{\ell=1}^{L},\,Ans
\Big),
\end{equation}
where $L$ denotes the number of reasoning steps, $rat_\ell^{\mathrm{T}}$ is a textual rationale,
$a_\ell$ is an atomic visual action, $rat_\ell^{\mathrm{M}}$ is the post-action visual rationale (an image),
and $Ans$ is the final textual answer.

\paragraph{Policy and execution.}
At each step $\ell$, the policy $\pi_\theta$ produces a textual rationale and selects an action:
\begin{equation}
\label{eq:policy_step}
\small
(rat_\ell^{\mathrm{T}},\,a_\ell) \sim \pi_\theta\!\left(\cdot \mid x^{\mathrm{T}},\,rat_{\ell-1}^{\mathrm{M}},\,\{(rat_j^{\mathrm{T}},a_j)\}_{j<\ell}\right).
\end{equation}
The selected action is then executed by a renderer/executor $\mathsf{R}$ that deterministically updates the visual state and
produces exactly one post-action image:
\begin{equation}
\label{eq:render_step}
\small
rat_\ell^{\mathrm{M}}=\mathsf{R}\!\left(rat_{\ell-1}^{\mathrm{M}},\,a_\ell\right),\quad \ell=1,\ldots,L,
\end{equation}
with initialization
\begin{equation}
\label{eq:render_init}
\small
rat_0^{\mathrm{M}} := x^{\mathrm{M}}.
\end{equation}
Finally, the answer is generated conditioned on the full interaction history, e.g.,
\begin{equation}
\label{eq:final_ans}
\small
Ans \sim \pi_\theta\!\left(\cdot \mid x^{\mathrm{T}},\,rat_L^{\mathrm{M}},\,\{(rat_\ell^{\mathrm{T}},a_\ell)\}_{\ell=1}^{L}\right).
\end{equation}

\paragraph{Atomic action space.}
We use a fixed set of atomic visual actions:
\begin{equation}
\label{eq:actions}
\small
\mathcal{A}=\{\textsf{ZOOM-in},\textsf{BBOX},\textsf{MARK},\textsf{LINE},\textsf{PRED}\},
\end{equation}
whose argument formats and semantics are specified next.

\subsection{Atomic Action Protocol Details}
\label{app:action-protocol}

This subsection specifies the atomic visual action vocabulary $\mathcal{A}$ used in the interleaved trajectories
(Eq.~\ref{eq:traj}--\ref{eq:actions}). Each action $a_\ell\in\mathcal{A}$ is applied to the current visual state
$rat_{\ell-1}^{\mathrm{M}}$ and yields exactly one post-action image $rat_{\ell}^{\mathrm{M}}$ via the executor $\mathsf{R}$
(Eq.~\ref{eq:render_step}).

\paragraph{Coordinate convention.}
All spatial arguments are in normalized image coordinates w.r.t.\ $rat_{\ell-1}^{\mathrm{M}}$:
$(0,0)$ is the top-left and $(1,1)$ is the bottom-right corner.
For boxes, we use $(x,y,w,h)$ for the top-left corner and width/height in $[0,1]$.
Out-of-range or degenerate arguments are treated as format errors.

\paragraph{\textsf{ZOOM-in}$(x,y,w,h)$.}
Produces $rat_{\ell}^{\mathrm{M}}$ by cropping the specified region from $rat_{\ell-1}^{\mathrm{M}}$
and resizing it to a fixed resolution (e.g., $512{\times}512$).

\paragraph{\textsf{BBOX}$(x,y,w,h)$.}
Produces $rat_{\ell}^{\mathrm{M}}$ by overlaying a bounding box on $rat_{\ell-1}^{\mathrm{M}}$
at the specified region.

\paragraph{\textsf{MARK}$(x,y,\mathrm{id})$.}
Produces $rat_{\ell}^{\mathrm{M}}$ by rendering a visual highlight on $rat_{\ell-1}^{\mathrm{M}}$
to draw attention to a target instance or region. The highlight may take various forms
(e.g., a colored semi-transparent overlay, halo, marker, or contour) anchored at $(x,y)$.
Optionally, a lightweight identifier $\mathrm{id}$ may be included to facilitate references in $rat_{\ell}^{\mathrm{T}}$.

\paragraph{\textsf{LINE}$(x_1,y_1,x_2,y_2)$.}
Produces $rat_{\ell}^{\mathrm{M}}$ by drawing a line between two normalized endpoints on $rat_{\ell-1}^{\mathrm{M}}$.

\paragraph{\textsf{PRED}$(\Delta)$.}
Produces $rat_{\ell}^{\mathrm{M}}$ by applying a one-step, task-dependent state transition to $rat_{\ell-1}^{\mathrm{M}}$,
where $\Delta$ is a serialized delta (text) specifying the update.
\emph{Note:} \textsf{PRED} updates the visual state (the post-action image) rather than directly outputting the final answer.

\section{Rule-based verifier for $\mathcal{R}_\mathrm{Acc}$}\label{app:verifier}
This appendix summarizes the rule-based verifier used to compute the accuracy reward $\mathcal{R}_\mathrm{Acc}$.
The verifier is deterministic and lightweight: it extracts a final answer from the generated trajectory and compares it to the ground truth using a small set of ordered checks.
The design emphasizes reliability (avoiding false positives) and robustness to superficial formatting variation.

\paragraph{Answer extraction and normalization.}
The verifier first locates a dedicated final-answer segment using an explicit marker (e.g., \texttt{Final Answer:}).
The extracted string is then lightly normalized to remove non-semantic variation such as whitespace and punctuation differences, and to strip common answer-introduction phrases (e.g., ``Answer is \dots'').
This step improves the stability of the evaluation without relaxing the core matching criteria.

\paragraph{Prioritized matching.}
The verifier applies checks in a reliability-first order:
\emph{(i) numeric matching} when both answers can be interpreted as numbers (including common formats such as percentages);
\emph{(ii) symbolic/mathematical equivalence} when the answer contains explicit mathematical expressions;
\emph{(iii) conservative textual matching} for general free-form strings;
and \emph{(iv) minimal domain-specific handling} for a small set of high-risk patterns (e.g., multiple-choice labels or chemistry-style strings) to reduce known false-positive cases.
When multiple checks are applicable, the verifier prefers the most reliable one.

\paragraph{Scoring.}
The accuracy reward $\mathcal{R}_\mathrm{Acc}\in[0,1]$ reflects the degree of match.
Checks based on numeric or symbolic equivalence typically yield binary outcomes, while textual matching may assign partial credit only when the match is sufficiently close under conservative criteria.
The format reward $\mathcal{R}_\mathrm{Fmt}$ is computed independently and only indicates whether the trajectory is well-formed and parsable.

\section{Dataset Details}
\label{sec:bench_des}

\paragraph{Training data.}
For Omni-R1, we sample Zebra-CoT~\citep{li2025zebra}, an 182K image-text interleaved multimodal reasoning dataset, as the supervised data for the PeSFT stage.
For Omni-R1-Zero, the PeSFT-stage data is bootstrapped from a small text-only CoT subset of M3CoT~\citep{chen-etal-2024-m3cot} using the synthetic interleaved construction in Section~\ref{subsec:omnir1zero}, yielding 791 interleaved samples.
For the PeRPO stage, we subsample approximately 10\% of Zebra-CoT and additionally include 1K ArxivQA~\citep{li-etal-2024-multimodal-arxiv} and 250 Geometry3K~\citep{lu2021inter} training samples, which do not provide interleaved multimodal reasoning annotations.

\paragraph{General benchmarks.}
We report detailed information on the general benchmarks, including task formulations, dataset sizes, and evaluation metrics. Detailed description is shown in Table~\ref{tab:general-bench-details}.

\begin{table*}[t]
\centering
\footnotesize
\setlength{\tabcolsep}{6pt}
\renewcommand{\arraystretch}{1.12}
\begin{minipage}{\textwidth}
\centering
\begin{tabular}{p{0.20\textwidth} p{0.74\textwidth}}
\toprule
Benchmark & Details \\
\midrule

MME-P~\citep{fu2023mme} &
\textbf{Task:} Binary visual QA focusing on perception and recognition.\newline
\textbf{Size:} 1{,}536 questions.\newline
\textbf{Metric:} MME score for the perception split.\newline
\textbf{Notes:} Reported as \textit{MME-P} in Table~\ref{tab:general_bench}. \\

\midrule
MME-R~\citep{fu2023mme} &
\textbf{Task:} Binary visual QA emphasizing reasoning and cognition.\newline
\textbf{Size:} 280 questions.\newline
\textbf{Metric:} MME score for the cognition split.\newline
\textbf{Notes:} Reported as \textit{MME-R} in Table~\ref{tab:general_bench}. \\

\midrule
MM-Vet~\citep{yu2024mmvet} &
\textbf{Task:} Open-ended multimodal QA covering a wide range of capabilities.\newline
\textbf{Size:} 218 examples.\newline
\textbf{Metric:} LLM-judge score on a 0--100 scale.\newline
\textbf{Notes:} We use the standard evaluation protocol in VLMEvalKit. \\

\midrule
V$^\ast$-Bench~\citep{wu2024vstar} &
\textbf{Task:} Detail-sensitive visual understanding and visual search.\newline
\textbf{Size:} 191 images.\newline
\textbf{Metric:} Accuracy.\newline
\textbf{Notes:} Reported as \textit{V$^\ast$} in Table~\ref{tab:general_bench}. \\

\midrule
POPE~\citep{li2023pope} &
\textbf{Task:} Yes/No probing for object hallucination.\newline
\textbf{Size:} 3{,}000 questions.\newline
\textbf{Metric:} Overall F1 score.\newline
\textbf{Notes:} In Table~\ref{tab:general_bench}, POPE corresponds to Overall F1 rather than accuracy. \\

\midrule
MMVP~\citep{tong2024mmvp} &
\textbf{Task:} Diagnostic visual patterns with challenging paired examples.\newline
\textbf{Size:} 300 questions.\newline
\textbf{Metric:} Accuracy.\newline
\textbf{Notes:} Evaluated with the default setup in VLMEvalKit. \\

\midrule
BLINK~\citep{fu2024blink} &
\textbf{Task:} Multi-choice evaluation of core visual perception skills across diverse sub-tasks.\newline
\textbf{Size:} 3{,}807 questions.\newline
\textbf{Metric:} Accuracy.\newline
\textbf{Notes:} Evaluated with the default setup in VLMEvalKit. \\

\bottomrule
\end{tabular}
\end{minipage}
\caption{Details of general multimodal benchmarks.}
\label{tab:general-bench-details}
\end{table*}

\section{Training Details}
\label{sec:appendix-hparams}

\paragraph{Implementation Details.}
Our implementation is built upon the open-source libraries verl~\citep{sheng2024hybridflow}, Transformers~\citep{wolf-etal-2020-transformers}, Flow-GRPO~\citep{liu2025flowgrpotrainingflowmatching} and trl~\citep{vonwerra2022trl}, following their respective licenses.

\paragraph{Hyperparameter Settings.}
We summarize the hyperparameter settings used in our experiments in Table~\ref{tab:hyperparams}. The configurations are organized into four groups: (i) Objective, (ii) Batching, (iii) Optimization, and (iv) Rollout.

\begin{table*}[t]
    \centering
    \small
    \setlength{\tabcolsep}{8pt}
    \renewcommand{\arraystretch}{1.15}
    \begin{tabular}{@{}lcl@{}}
        \toprule
        \textbf{Hyperparameter} 
        & \textbf{Setting} 
        & \textbf{Description} \\
        \midrule

        \multicolumn{3}{@{}l}{\textbf{Objective}} \\
        \midrule
        KL Loss Coefficient          & 0.003            & KL regularization weight \\
        PPO Clip Range (Low)         & 0.20             & lower clipping bound \\
        PPO Clip Range (High)        & 0.28             & upper clipping bound \\
        \midrule

        \multicolumn{3}{@{}l}{\textbf{Batching}} \\
        \midrule
        Train Batch Size   & 64               & prompts per update \\
        Gen Batch Size     & 192              & $3\times$ train batch \\
        PPO Mini-batch Size          & 16               & per PPO mini-batch \\
        \midrule

        \multicolumn{3}{@{}l}{\textbf{Optimization}} \\
        \midrule
        Learning Rate                & $1\times10^{-7}$ & peak LR \\
        Warmup Steps                 & 5                & LR warmup \\
        Weight Decay                 & 0.10             & L2 regularization \\
        Entropy Coefficient          & 0.001            & entropy bonus weight \\
        Gradient Clipping     & 1.0              & global-norm clip \\
        Loss Aggregation             & seq-mean-token-mean & loss reduction \\
        \midrule

        \multicolumn{3}{@{}l}{\textbf{Rollout}} \\
        \midrule
        Responses per Prompt         & 16               & samples per prompt \\
        Temperature                  & 1.0              & sampling temperature \\
        Top-$p$                      & 0.95             & nucleus threshold \\
        \bottomrule
    \end{tabular}
    \caption{Training Hyperparameters. Settings used in our experiments.}
    \label{tab:hyperparams}
\end{table*}

\section{Additional Results}
\label{section:additional_res}
\begin{figure*}[t]
    \centering
    \includegraphics[width=0.7\linewidth]{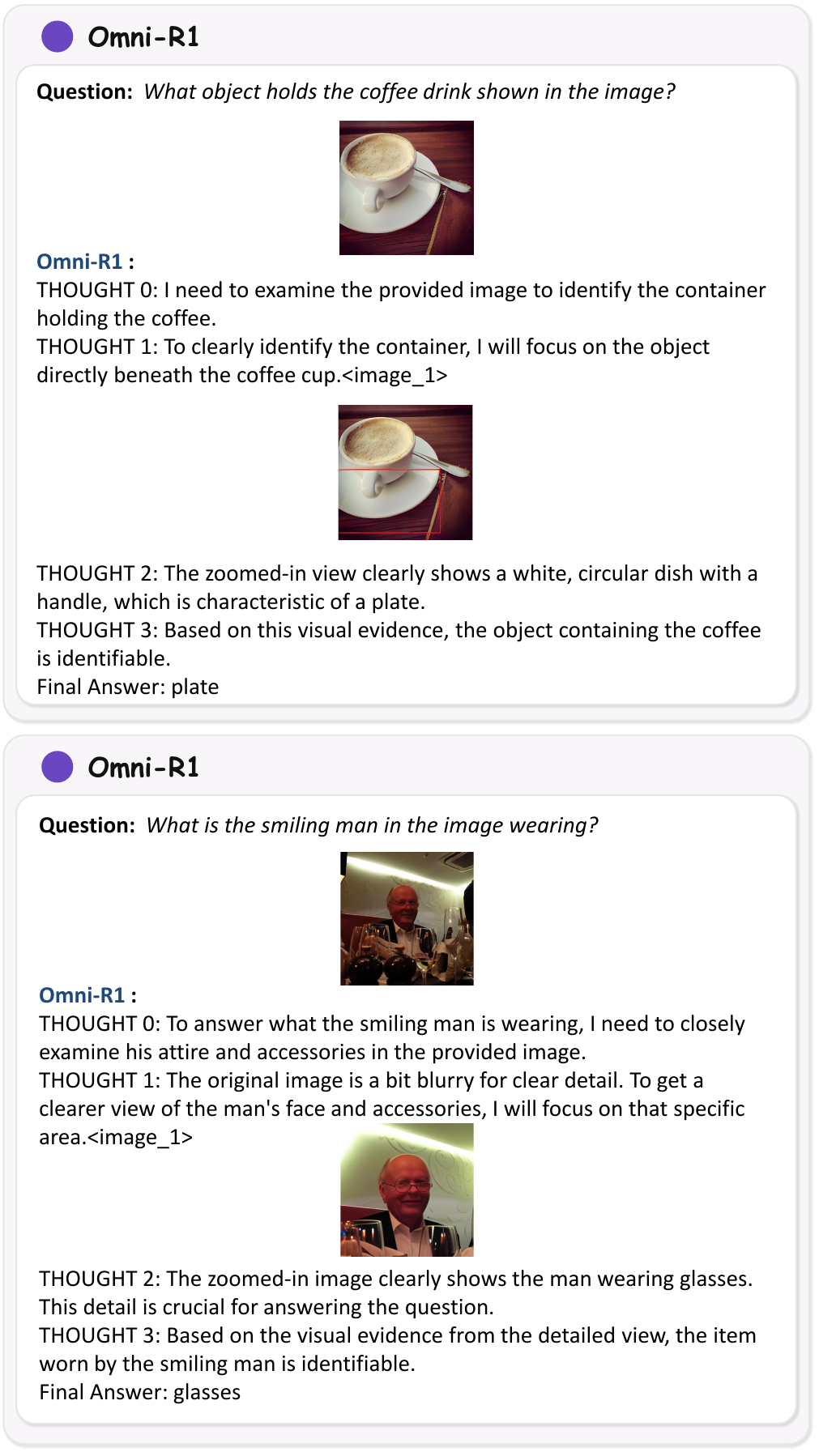}
    \caption{Additional case results of Omni-R1.}
    \label{fig:additional_case1_omni_r1}
\end{figure*}

\begin{figure*}[t]
    \centering
    \includegraphics[width=0.7\linewidth]{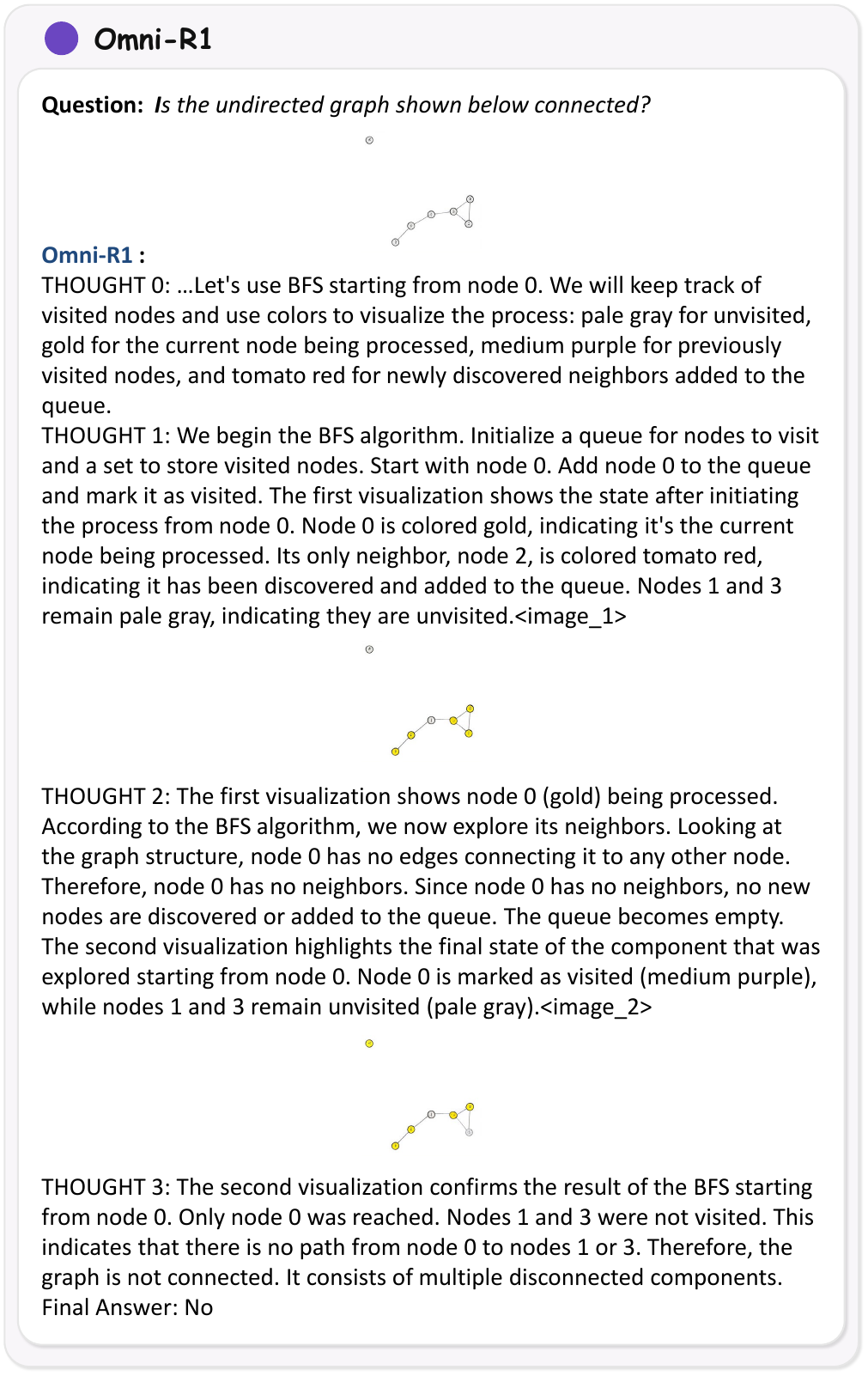}
    \caption{Additional case results of Omni-R1.}
    \label{fig:additional_case2_omni_r1}
\end{figure*}

\begin{figure*}[t]
    \centering
    \includegraphics[width=0.7\linewidth]{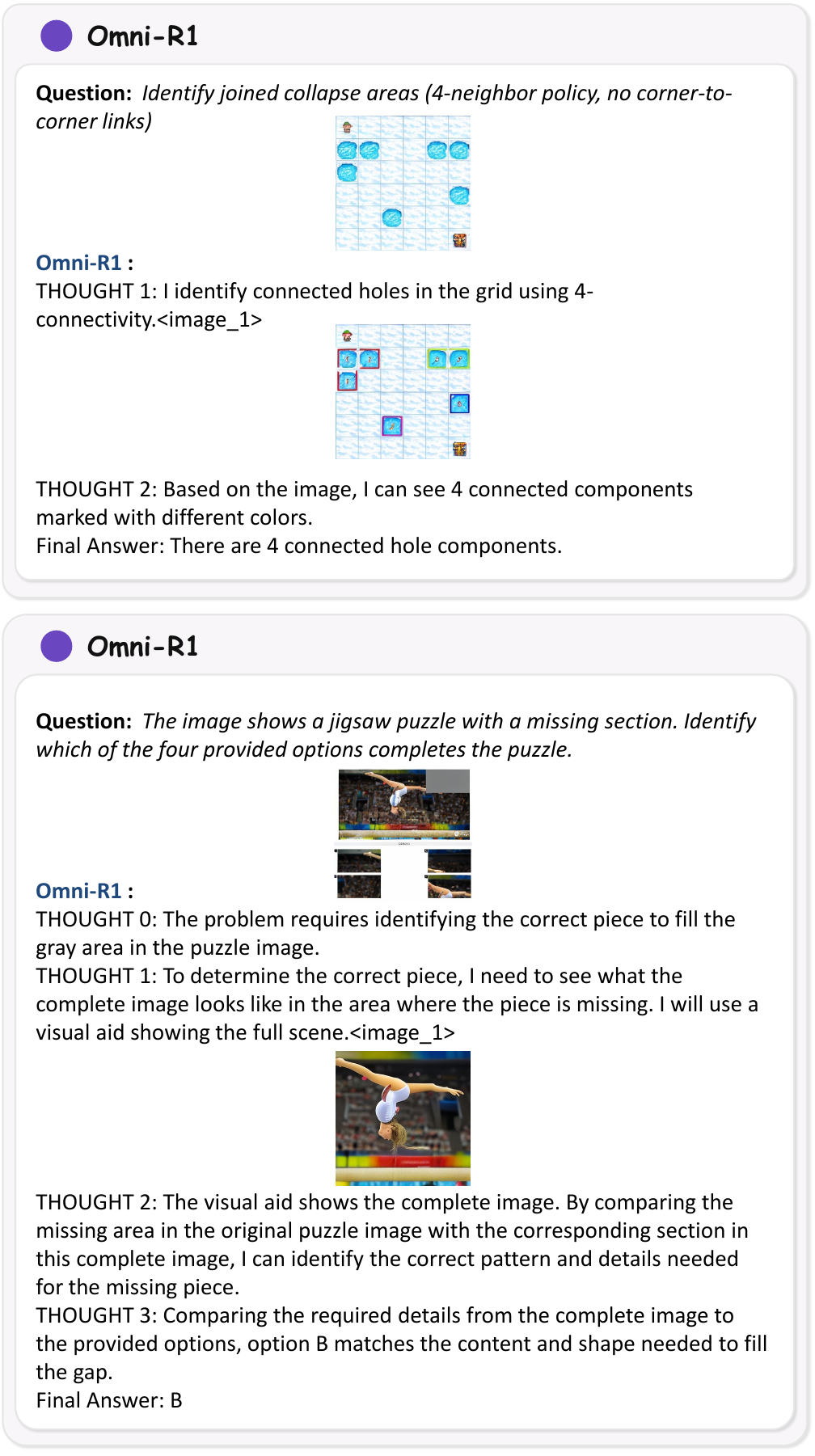}
    \caption{Additional case results of Omni-R1.}
    \label{fig:additional_case3_omni_r1}
\end{figure*}

To qualitatively illustrate how Omni-R1 performs generative multimodal reasoning, we present additional full interleaved trajectories in Figures~\ref{fig:additional_case1_omni_r1}--\ref{fig:additional_case3_omni_r1}. These examples span diverse tasks and capability requirements, showing how the model generates intermediate visual evidence and composes Uni-Skills across different scenarios.

\section{Prompt Templates}
\label{section:prompt_templates}
To reduce evaluation variance and ensure reproducibility on Omni-Bench, we adopt a fixed LLM-judge prompt template, shown in Table~\ref{prompt_gpt_binary_judge}. The template enforces a binary decision using only the ground-truth answer and the model output, and standardizes the handling of common edge cases, including numeric formatting, unit conversion, and multiple-choice variants.

\begin{table*}[!htbp]
    \small
    \centering
    \setlength{\tabcolsep}{5pt}
    \renewcommand{\arraystretch}{1.2}
    \begin{tabularx}{\textwidth}{X}
        \toprule
        \textbf{System Prompt}:\\
You are a strict QA judge.\\
Decide correctness by comparing ONLY the ground-truth answer and the model answer. The question may be multiple-choice or open-ended.\\
\\
OUTPUT:\\
Return exactly one token with NO quotes/punctuation/spaces/code fences: True or False.\\
\\
GENERAL RULES:\\
1) Judge factual/semantic equivalence; ignore phrasing, filler, or reasoning text. If the model's final claim contradicts the ground truth or hedges without committing, return False.\\
2) Numbers: allow formatting differences (1,000 vs 1000), scientific notation, or rounding that preserves the stated value. If units are present, require the same value after conversion; missing/extra incompatible units => False.\\
3) Lists/sets: require the same items; order doesn't matter. Missing or extra items => False.\\
4) Spans/names: accept common synonyms and aliases that uniquely indicate the same entity.\\
5) If ambiguous, empty, multiple conflicting answers, or cannot be judged, return False.\\
\\
SPECIAL RULES FOR MULTIPLE-CHOICE (only when options are provided below):\\
A) Treat option LETTERS and their NUMERIC ORDINALS as equivalent (A=1, B=2, C=3, ...), but ONLY within this question's options.\\
B) Treat the CORRECT OPTION'S FULL TEXT as equivalent to its letter and numeric index.\\
\\
Your final output must be only the single token: True or False.\\
        \midrule
        \textit{\textbf{User Prompt}}:\\
\textbf{\#\#\# Ground Truth}\\
\{Ground Truth\}\\
\\
\textbf{\#\#\# Model Answer}\\
\{Model Answer\}\\
\\
\textbf{\#\#\# Decision}\\
Return only one of: True, False.\\
        \bottomrule
    \end{tabularx}
\caption{Template for Omni-Bench's GPT Judge.}
\label{prompt_gpt_binary_judge}
\end{table*}

\end{document}